# Token-Hungry, Yet Precise: DeepSeek R1 Highlights the Need for Multi-Step Reasoning Over Speed in MATH.

Evgenii Evstafev [A]

[A] University Information Services (UIS), University of Cambridge,
Roger Needham Building, 7 JJ Thomson Ave, Cambridge CB3 0RB, UK, ee345@cam.ac.uk

**ABSTRACT**

*This study investigates the performance of the DeepSeek R1 language model on 30 challenging mathematical problems derived from the MATH dataset, problems that previously proved unsolvable by other models under time constraints. Unlike prior work, this research removes time limitations to explore whether DeepSeek R1's architecture, known for its reliance on token-based reasoning, can achieve accurate solutions through a multi-step process. The study compares DeepSeek R1 with four other models (gemini-1.5-flash-8b, gpt-4o-mini-2024-07-18, llama3.1:8b, and mistral-8b-latest) across 11 temperature settings. Results demonstrate that DeepSeek R1 achieves superior accuracy on these complex problems but generates significantly more tokens than other models, confirming its token-intensive approach. The findings highlight a trade-off between accuracy and efficiency in mathematical problem-solving with large language models: while DeepSeek R1 excels in accuracy, its reliance on extensive token generation may not be optimal for applications requiring rapid responses. The study underscores the importance of considering task-specific requirements when selecting an LLM and emphasizes the role of temperature settings in optimizing performance.*

**TYPE OF PAPER AND KEYWORDS**

*Empirical Research Paper, Large Language Models, Mathematical Reasoning, DeepSeek R1, Tokenization, Accuracy, Efficiency, Temperature, MATH Dataset.*

## 1 INTRODUCTION

This study investigates the DeepSeek R1 [1] model's performance on a challenging subset of mathematical problems derived from the MATH dataset [2]. Prior research demonstrated that these specific problems remained unsolved by several language models when operating under strict time constraints [3]. Unlike those previous experiments, this research removes time limitations to explore the hypothesis that DeepSeek R1's architecture, with its documented reliance on token-based reasoning, facilitates accurate solutions through a more deliberate, multi-step process. The central aim is to analyze the relationship between a model's accuracy in solving these complex mathematical problems and its token usage, providing insights into the computational cost associated with achieving high accuracy in this domain.

## 2. BACKGROUND AND RELATED WORK

Recent advancements in natural language processing have led to increased interest in applying large language models (LLMs) to mathematical problem-solving. While initial symbolic approaches faced limitations in handling natural language nuances, the advent of transformer-based models [4] significantly improved the ability of LLMs to process and generate mathematical text. Consequently, models have been developed that can solve basic mathematical word problems [5]. However, more complex problems, such as those found in the MATH dataset [6], which often require multi-step reasoning and symbolic manipulation, continue to pose a significant challenge. Benchmarks using the MATH dataset have revealed that even state-of-the-art models struggle to achieve high accuracy, particularly under resource constraints [3]. The DeepSeek R1 model, the focus of this study, is of particular interest due to its documented reliance on token-based reasoning steps, suggesting a potential mechanism for enhanced accuracy through a more iterative process. Furthermore, the influence of temperature settings on model outputs, affecting the balance between creativity and coherence, warrants consideration in the context of mathematical reasoning. This research builds upon prior work by investigating the interplay between model architecture, resource utilization (specifically token generation), and the impact of temperature on the ability of DeepSeek R1 and other leading LLMs to solve challenging mathematical problems.

## 3. METHODOLOGY

This study builds upon the findings of a previous benchmark experiment, "Token-by-Token Regeneration and Domain Biases: A Benchmark of LLMs on Advanced Mathematical Problem-Solving" [3], which evaluated the performance of various large language models (LLMs) on the MATH dataset. The previous study imposed strict time limits on response generation to prevent infinite loops, a constraint that significantly hindered the performance of the DeepSeek R1 model. This current experiment aims to explore the capabilities of DeepSeek R1 and other LLMs on a subset of the MATH dataset without these time limitations, focusing instead on identifying and mitigating repetitive response patterns.

### 3.1 DATASET CREATION

The dataset for this study was derived from the results of the aforementioned previous experiment. Specifically, 30 problems from the MATH dataset were selected, identified as problems that no model in the original study was able to solve correctly within the imposed time constraints. These 30 problems constitute the dataset for this experiment, representing a challenging subset of mathematical reasoning tasks.

### 3.2 MODEL SELECTION

Five distinct LLMs were selected for evaluation in this experiment:

1. deepseek-r1:8b [1]: This model is the primary focus of this study, as it was significantly impacted by the time constraints in the previous experiment. Official documentation suggests optimal performance with temperature settings between 0.6 and 0.8 and cautions against the use of system prompts.
2. gemini-1.5-flash-8b [7]: A model from Google, representing a different architectural approach.
3. gpt-4o-mini-2024-07-18 [8]: A recent model from OpenAI, included for comparative analysis.
4. llama3.1:8b [9]: An open-source model known for its performance on various benchmarks.
5. mistral-8b-latest [10]: Another strong open-source model, serving as a point of comparison.

Each model was tested with 11 different temperature settings, ranging from 0.0 to 1.0 with increments of 0.1. This resulted in a total of 1650 experimental runs (30 problems, 5 models, 11 temperatures).

### 3.3 EVALUATION METRICS

The primary evaluation metric was the correctness of the solution to each problem. Due to the potential for overly verbose or repetitive responses, the following procedure was implemented:

- Responses exceeding 1000 characters were truncated, retaining only the final 1000 characters for evaluation. This assumes that the final portion of the response contains the model's ultimate answer.
- To address potential infinite loops, a repetition detection mechanism was implemented. If the last 40 characters of the response were found to be repeated 400 times, the generation process was forcibly terminated.
- The final or truncated response was compared against the known correct answer. The correctness of each response was evaluated using a binary metric: 1 if the model's answer matched the expected answer exactly, and 0 otherwise. This evaluation was performed using the mistral-large-2411 [11] model as a judge.

In addition to correctness, the average number of tokens generated per model across all successful runs was calculated.

## 4. RESULTS

The experiment yielded 1650 data points, representing the responses of each model to each problem at each temperature setting. Only a subset of these runs resulted in correct answers. Table 1 presents the average token count for each model across their respective successful runs. Figure 1 visually depicts these results, highlighting the significant difference in token usage between DeepSeek R1 and the other models.

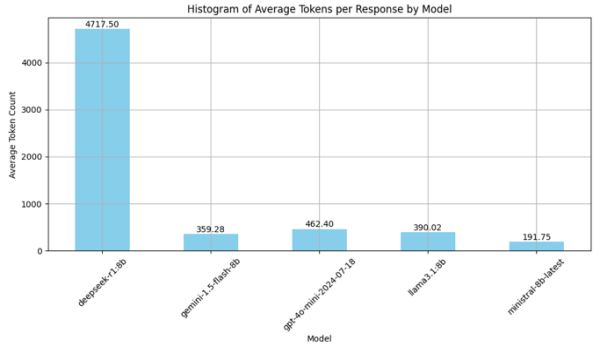

**Figure 1: Average Token Count per Model (Successful Runs Only)**

Token-by-token regeneration improved Algebra performance but showed neutral or negative effects in other domains.

**Table 1: Average Token Count per Model**

| Model | Average Tokens |
|---|---|
| deepseek-r1:8b | 4717.500000 |
| gemini-1.5-flash-8b | 359.283333 |
| gpt-4o-mini-2024-07-18 | 462.398268 |
| llama3.1:8b | 390.020000 |
| mistral-8b-latest | 191.750000 |

## 5. KEY OBSERVATIONS

The results [12] clearly demonstrate that DeepSeek R1, while capable of solving complex mathematical problems that eluded other models in the previous constrained experiment, does so at the cost of significantly higher token usage. The average token count for DeepSeek R1 (4717.5) is an order of magnitude higher than that of the other models tested. This observation aligns with the model's architectural design, which, according to its documentation, relies heavily on token-based reasoning steps, even implying a requirement for these "reasoning tokens" for proper functioning.

Furthermore, the experiment underscores the importance of temperature settings in influencing model behavior. The observation that Llama 3.1 only achieved correct results at a temperature of 0.4 highlights the sensitivity of certain models to this parameter and suggests that optimal performance may require fine-tuning beyond the default settings.

These findings suggest a trade-off between speed and accuracy in the context of complex mathematical problem-solving. While DeepSeek R1 demonstrates superior accuracy on challenging problems when given sufficient computational resources (in terms of token generation), its performance comes at the expense of significantly longer processing times compared to models that produce more concise, albeit potentially less accurate, responses. This highlights the need for careful consideration of the specific requirements of a given task when selecting an appropriate LLM. For tasks requiring rapid responses, a model like Mistral might be preferable, whereas tasks prioritizing accuracy on complex problems might benefit from DeepSeek R1's more deliberate, token-intensive approach.

Further research should explore the internal mechanisms of DeepSeek R1 to better understand the role of "reasoning tokens" and to investigate potential optimizations that could reduce token usage without sacrificing accuracy. Additionally, exploring the impact of different prompt engineering strategies on model performance, particularly for models like DeepSeek R1, could provide valuable insights into maximizing their capabilities.

## 6 SUMMARY AND CONCLUSIONS

This study examined the performance of five large language models on 30 challenging mathematical problems, specifically focusing on the DeepSeek R1 model's ability to solve problems that were unsolvable by other models under previous time constraints. The results demonstrate that DeepSeek R1 can achieve high accuracy on these complex problems when allowed to generate a significantly larger number of tokens, confirming its reliance on a multi-step reasoning process. However, this approach results in a much higher token count compared to other models, indicating a trade-off between accuracy and efficiency.

In conclusion, the findings highlight the importance of considering the specific requirements of a task when selecting a large language model for mathematical problem-solving. While DeepSeek R1 excels in accuracy on challenging problems, its token-intensive approach may not be suitable for applications requiring rapid responses. Conversely, models that generate fewer tokens may be faster but less accurate on complex tasks. The research underscores the need for a nuanced understanding of the strengths and limitations of different LLM architectures and emphasizes the significant role of factors like temperature settings in optimizing performance.

**AUTHOR BIOGRAPHIES**

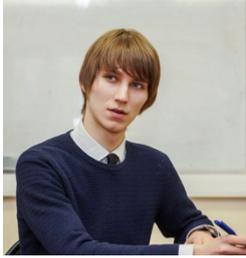 Evgenii Evstafev is a skilled software developer at the University of Cambridge, where he has been working since September 2022, specializing in identity and access management. He earned a Bachelor's degree in Business Informatics from the Higher School of Economics (2010-2014) and a Master's degree in Programming from Perm National Research Polytechnic University (2014-2016). Evgenii also taught at the Faculty of Mechanics and Mathematics at Perm State University while engaged in postgraduate studies (2016-2019). His professional journey spans over 11 years across various industries, including roles such as System Architect at L'Etoile (2021-2022) focusing on product development, the Head of Analytics at Gazprombank (2020-2021), and Head of the Department for System Analysis and Software Design at Center 2M (2019-2020). Additionally, he worked on system development at the energy company T Plus (2016-2019).